\documentclass[10pt, a4paper]{article}

\usepackage{lrec-coling2024} 
\usepackage{times}
\usepackage{latexsym}

\usepackage{blindtext}
\usepackage{multirow}
\usepackage{bm}
\usepackage{bbm} 
\usepackage{comment}
\usepackage{subcaption}
\usepackage{array} 
\usepackage{booktabs}
\usepackage{wrapfig}
\usepackage[safe]{tipa} 
\usepackage{color, colortbl}
\usepackage[misc,geometry]{ifsym} 
\usepackage{graphicx}
\usepackage{amsmath}
\usepackage{amssymb}
\usepackage{mathrsfs}
\usepackage{booktabs}
\usepackage[ruled,vlined]{algorithm2e}
\usepackage{algorithmic}

\usepackage[T1]{fontenc}

\usepackage[utf8]{inputenc}

\usepackage{microtype}

\usepackage{inconsolata}

\newcommand{\tableCellHeight}{1}

\newcommand{\tabstyle}[1]{
  \setlength{\tabcolsep}{#1}
  \renewcommand{\arraystretch}{\tableCellHeight}
  \centering
  \small
}

\title{Soft-Prompting with Graph-of-Thought for Multi-modal Representation Learning}

\name{Juncheng Yang\textsuperscript{\rm 1,}\textsuperscript{\rm 3,}$^{*}$, Zuchao Li\textsuperscript{\rm 1,}\textsuperscript{\rm 2,}$^{\dag}$, Shuai Xie\textsuperscript{\rm 4,}$^{*}$, Wei Yu\textsuperscript{\rm 1}, Shijun Li\textsuperscript{\rm 1,}$^{\dag}$, Bo Du\textsuperscript{\rm 1,}\textsuperscript{\rm 2}} 
\address{\textsuperscript{\rm 1}School of Computer Science, Wuhan University, Wuhan, Hubei, China \\
\textsuperscript{\rm 2}National Engineering Research Center for Multimedia Software, Wuhan, Hubei, China\\
\textsuperscript{\rm 3}School of Electronic Information Engineering, Henan Polytechnic Institute, Nanyang, Henan, China\\
\textsuperscript{\rm 4}JD Explore Academy, Beijing, China \\
\normalsize{\texttt{\{yjuncheng,zcli-charlie,yuwei,shjli,dubo\}@whu.edu.cn}, \texttt{xieshuai@jd.com}} \\
}

\abstract{
The chain-of-thought technique has been received well in multi-modal tasks. It is a step-by-step linear reasoning process that adjusts the length of the chain to improve the performance of generated prompts. However, human thought processes are predominantly non-linear, as they encompass multiple aspects simultaneously and employ dynamic adjustment and updating mechanisms. Therefore, we propose a novel Aggregation-Graph-of-Thought (AGoT) mechanism for soft-prompt tuning in multi-modal representation learning. The proposed AGoT models the human thought process not only as a chain but also models each step as a reasoning aggregation graph to cope with the overlooked multiple aspects of thinking in single-step reasoning. This turns the entire reasoning process into prompt aggregation and prompt flow operations. Experiments show that our multi-modal model enhanced with AGoT soft-prompting achieves good results in several tasks such as text-image retrieval, visual question answering, and image recognition. In addition, we demonstrate that it has good domain generalization performance due to better reasoning.
 \\ \newline  }

\begin{document}

\maketitleabstract

\renewcommand{\thefootnote}{}
\footnotetext{$^{*}$Equal Contribution.}
\footnotetext{$^{\dag}$Correspondence to: Zuchao Li, Shijun Li.}

\section{Introduction}

Multi-modal representation learning includes the fusion of different modal data such as visual, text, and sound which is essential for multi-modal tasks like text-to-speech synthesis~\cite{ren2019fastspeech}, audio-visual-speech recognition~\cite{afouras2018deep}, text-image retrieval~\cite{gabeur2022masking,yu2022u}, vision-language recognition~\cite{zhou2022conditional}, and visual question answering~\cite{li2022align,zheng2021knowledge}.%

In vision-language models, prompt learning has been widely adopted as an effective strategy~\cite{zhu2023prompt,ge2023chain}.%
The prompt was first applied in the field of NLP, and its role is to give a prompt to the pre-trained language model to help it understand human problems better. The earliest prompts were directly spliced with the original text into discrete characters that remained unchanged during the training process, known as hard prompts~\cite{DBLP:conf/eacl/SchickS21, DBLP:journals/aiopen/HanZDLS22}. This was later developed into soft prompts~\cite{DBLP:conf/acl/GuHLH22, liu2022p}, which can be tuned based on contextual semantics during training. While in multi-modal tasks, CoOp~\cite{zhou2022learning} first turns a fixed token prompt in CLIP~\cite{radford2021learning} into a learnable token embedding. CoCoOp~\cite{zhou2022conditional} adds an instance image feature to CoOp, and KgCoOp~\cite{yao2023visual} uses a fixed text prompt and a learnable prompt as supervised signals to constrain the training of the CLIP model. However, these approaches rely on a single language cue to guide visual understanding but fail to be aware of a basic but very important feature of language, reasoning, which allows multi-modal models to quickly adapt to new scenarios.
\begin{figure}[t]
   \center{\includegraphics[width=8cm]{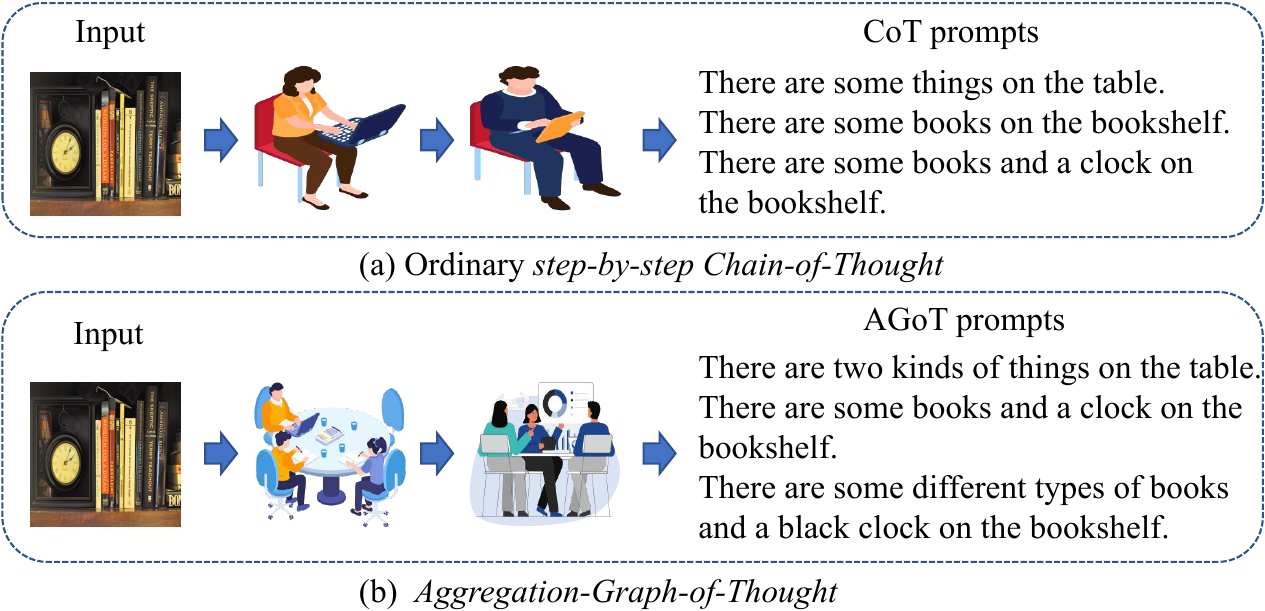}} 
   \caption{Comparison between (a) ordinary step-by-step Chain-of-Thought and (b) Aggregation-Graph-of-Thought.}
   \label{fig:introduction}
   \vspace{-1.2em}
\end{figure}

The Chain-of-Thought (CoT) technique simulates the reasoning process of the brain and imitates its behavior to improve the generalization of the model. For complex problems, instead of providing the answer directly, a string of reasoning sentences is used to generate the answer.
As shown in Figure~\ref{fig:introduction} (a), a picture containing some books and a clock is given by CoT, and then reason about the content of the picture step by step. From the initial description of ``There are some things on the table'' to ``There are some books on the bookshelf'', we can see that the statement's content becomes fuller. Finally, the phrase ``There are some books and a clock on the bookshelf'' completes a relatively complete description of the image. This step-by-step reasoning improves the model's visual understanding performance. CoT-PT~\cite{ge2023chain} was the first to adapt CoT to the field of vision-language tasks, combining visual and CoT cues to tune the model and achieve good results in downstream tasks.

Although the CoT technique considers the progressively deeper understanding process during reasoning, it fails to fully leverage the benefits of reasoning for multi-modal comprehension due to its disregard of the fact that comprehension can occur from multiple perspectives. In response to this characteristic, we propose a new CoT reasoning method called Aggregation-Graph-of-Thought (AGoT). As shown in Figure~\ref{fig:introduction} (b), when inputting a picture, each step in AGoT, from ``There are two kinds of things on the table'' to ``There are some books and a clock on the bookshelf'' to ``There are some different types of books and a black clock on the bookshelf'', each step of the reasoning process is more reasonable than Figure~\ref{fig:introduction} (a) due to understanding aggregation of multiple aspects. It's apparent that at each reasoning node, AGoT aggregates multiple levels of aspects, whereas CoT prompts consider only one level. From the simulated results on the far right, it can be observed that AGoT prompts, after aggregating multi-level aspects, provide a more comprehensive understanding of the problem. Specifically, the AGoT models each step of reasoning in the CoT as a reasoning aggregation graph to cope with the multiple-view thinking problem that is overlooked in single-step reasoning. Each aggregation node in the graph also serves as a node within the CoT. Multiple nodes are interconnected to form a chain, with each node's representation aggregated from the previous node, multiple subnodes, and vision features. The resultant node representation is then passed to the next one. We design a flow controller to control the degree of information flow between the previous node and the current one.

We used the strong baseline model CLIP as our base model and conducted experiments in vision-language understanding tasks, text-image retrieval (Flickr30k~\cite{plummer2015flickr30k}, MSCOCO~\cite{chen2015microsoft}), and visual question answering (VQAv2~\cite{DBLP:conf/iclr/ShenLTBRCYK22}). Our AGoT method achieved better results compared to CLIP +5.70\%, +5.40\%, +19.91\% in each of these tasks, and also showed improvement compared to CLIP + CoT, +1.70\%, +0.80\%, +0.88\% respectively. Additionally, since our AGoT can be used as a soft-prompting method, it can improve the generalization ability of visual classification tasks. We evaluated it on various visual classification tasks and found that AGoT achieved better domain generalization ability than the previous state-of-the-art method, demonstrating the effectiveness of the AGoT method.

In summary, our contributions are three-fold:
\begin{itemize}
    \item we propose a novel Aggregation-Graph-of-Thought mechanism for soft-prompt tuning in multi-modal representation learning.
    \item To cope with the overlooked multiple aspects of thinking in single-step reasoning, we model the reasoning step as an aggregation graph, and turn the whole process into a prompt aggregation and prompt flow operation.  
    \item AGoT exhibits strong multi-modal representation learning ability and achieves good results on 18 datasets such as text-image retrieval, VQA, and image classification and has good domain generalization ability.
\end{itemize} 

\begin{figure*}[t]
   \center{\includegraphics[width=16cm]{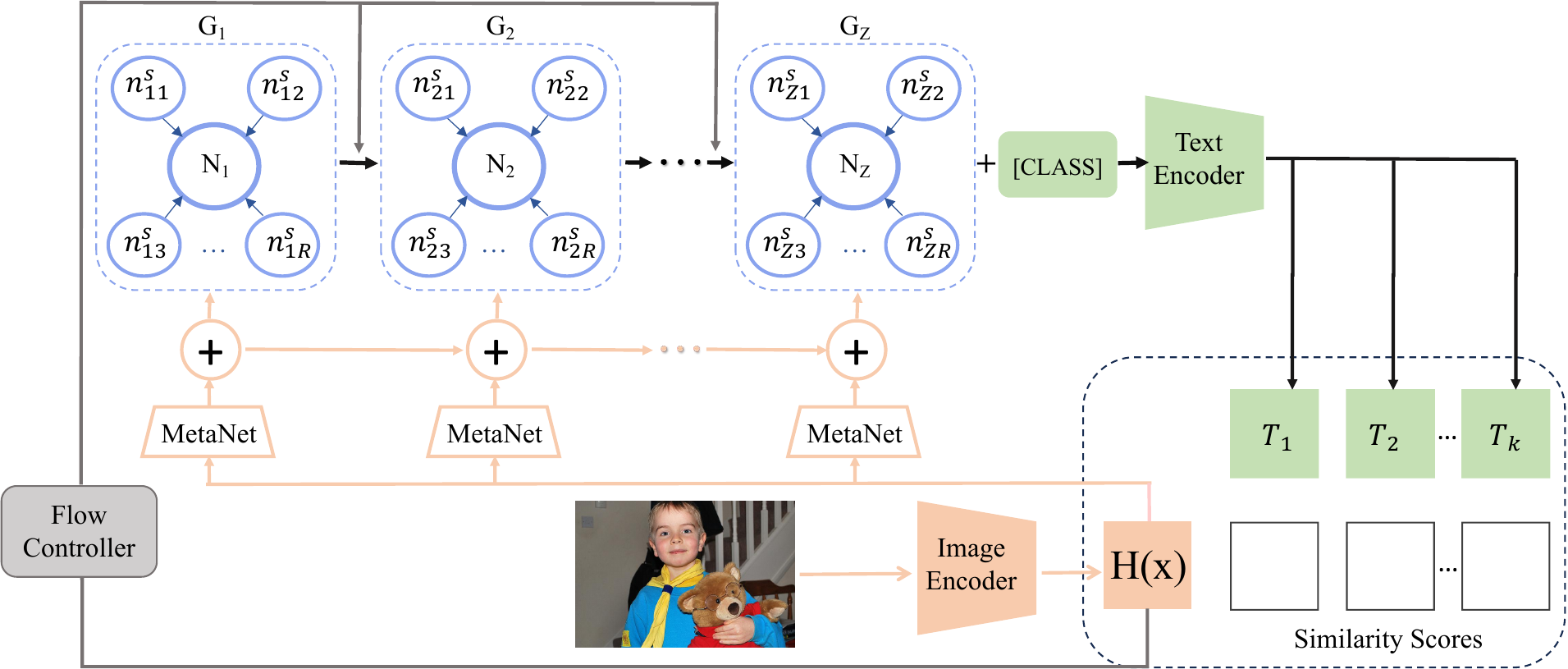}} 
   \caption{Illustration of the proposed AGoT. It learns a high-quality soft prompt with prompt aggregation ({\color{blue}blue}) and prompt flow operation ({\color{gray}gray}) for multi-view thinking and adaptation to complex multi-modal tasks.}
   \label{fig:framework}
   \vspace{-1.2em}
\end{figure*}

\section{Related Work}
\paragraph{Multi-modal Representation Learning}

Multi-modal representation learning has a significant number of models based on the structure of vision-language encoding. CLIP is a text encoder that uses Transformer~\cite{vaswani2017attention}, while the image encoder is either ResNet~\cite{he2016deep} or Vision Transformer~\cite{DBLP:conf/iclr/DosovitskiyB0WZ21}. The model is trained on 400 million image-text pairs and performs well on 30 datasets. ALIGN~\cite{jia2021scaling} is another typical vision-language model that uses contrast learning to train more than one billion noisy image-text pairs. It achieves state-of-the-art performance on the MSCOCO and Flickr30k datasets.

These multi-modal foundational models are widely used for downstream tasks, such as image classification, video-text retrieval, tracking, image captioning, and point cloud understanding.
CoOp enhances the generalization performance of the model by converting the fixed text description of \textit{``a photo of a''} in CLIP into a learnable token embedding. To address the challenge of poor performance on unseen data, CoCoOp incorporates instance features into the text description, resulting in further improvements in the model's performance. KgCoOp enhances the model's generalization by minimizing the disparity between learnable prompts and hand-crafted prompts. DualCoOp~\cite{sun2022dualcoop} proposes using dual context optimization as a unified framework to address the multi-label recognition tasks. ProGrad~\cite{zhu2023prompt} updates the prompt in a way that aligns its gradient with general knowledge, while ProDA~\cite{lu2022prompt} focuses on learning prompt prior distributions. 
These methods utilize language cues to assist in visual tasks. However, they fall short in replicating the innate reasoning process employed by humans, thereby missing out on the advantages of deep visual understanding acquired through vision-language reasoning.

\paragraph{Chain-of-Thought}
CoT technology excels in Large Language Model (LLM) few-shot reasoning and fine-tuning. Wei et al.~\cite{wei2022chain} was the first to use CoT as a discrete cue learning in LLM, with good results. Self-Consistency-CoT~\cite{DBLP:conf/iclr/0002WSLCNCZ23} improves the performance of the CoT approach significantly by using majority voting on answers. STaR~\cite{zelikman2022star} proposes a boosting approach that allows even small and medium-sized models to use CoT. AutoCoT~\cite{DBLP:conf/iclr/0001Z0S23} samples diverse problems and automatically generates reasoning chains. Shi et al.~\cite{DBLP:conf/iclr/ShiSF0SVCTRZ0W23} evaluate the strong multilingual reasoning power of CoT in solving multilingual grade school math problems. ToT~\cite{yao2024tree} allows LLM to choose a course of action by considering many different paths of reasoning and self-assessment.
CoT also has a wide range of applications in multi-modal tasks. Multimodal-CoT~\cite{zhang2023multimodal} integrates language and visual modalities into a two-stage framework, using multi-modal information to infer answers better. He et al.~\cite{he2023multi} propose generating effective image features consistent with language thoughts through the potential space diffusion process. CoT-PT~\cite{ge2023chain} uses CoT prompt tuning for vision-language models. VCoT~\cite{rose2023visual} incorporates visual augmentation into reasoning tasks, using CoT cues and visuals to bridge logical gaps in sequential data recursively. However, this method is geared towards generative tasks, while our method addresses downstream discriminative tasks.

\section{Methodology}
\subsection{Contrastive Learning}

To perform multi-modal representation learning, this work follows contrastive learning in paired text and image. Consistent with CLIP, let $x$ be the input image, and $t$ be the text sequence formed by the prompt and corresponding class. The model computes the probability of each class $c_k$:
\begin{equation}
  p(c_k|x) = \frac {e^{cos(\mathcal{G}(t_k), \mathcal{H}(x)) / \tau}}{\sum_{i=1}^K e^{cos(\mathcal{G}(t_i), \mathcal{H}(x))/ \tau}},
\end{equation}
where $cos(\cdot,\cdot)$ denotes the cosine similarity, $\mathcal{H}(x)$ stands for the image representation from the image encoder, $\mathcal{G}(t_k)$ stands for the text representation from the text encoder of concatenated prompt and class description sentences $t_k$, $K$ is the total number of classes, and $\tau$ is the temperature parameter.
The goal of contrastive learning is to learn a generalized image encoder $\mathcal{H}(\cdot)$ and text encoder $\mathcal{G}(\cdot)$ using limited pairs of data, so that the model has a cross-modal semantic space and can better transfer to downstream tasks such as text-image retrieval, VQA, and fine-grained classification.

\subsection{Prompt Learning and CoT}

CoOp is the first to turn the fixed prompt, such as \textit{``a photo of a''}, in the CLIP model into a learnable embedding $h^p$, thus soft-prompting. This overcomes the need for skill and engineering experience in manually creating prompts and surpasses CLIP in overall performance across 14 datasets for image recognition, fine-grained classification, and more. The probability for each class can be expressed as:
\begin{equation}
  p(c_k|x) = \frac {e^{cos(\mathcal{G}([h^p_k, t_k]), \mathcal{H}(x)) / \tau}}{\sum_{i=1}^K e^{cos(\mathcal{G}([h^p_i, t_i]), \mathcal{H}(x))/ \tau}},
\end{equation}
where $[h^p_k, t_k]$ denotes the concatenation of learnable prompt embedding and text embedding.

To solve the problem that CoOp performs poorly on unseen classes, CoCoOp further uses a simple \textit{MetaNet} network to generate conditional tokens for each image and then combines this image-side tokens with the text prompt, which dynamically adapts to the changes of each image instance, overcoming the problem of class shift. The prediction probability is computed as follows:
\begin{equation}
  p(c_k|x) = \frac {e^{cos(\mathcal{G}([h^p_k, M(x), t_k]), \mathcal{H}(x)) / \tau}}{\sum_{i=1}^K e^{cos(\mathcal{G}([h^p_i, M(x), t_i]), \mathcal{H}(x))/ \tau}},
\end{equation}
where $M(x)$ is the representation of conditional tokens from \textit{MetaNet}.

CoT-PT introduces CoT for prompt learning in vision-language tasks. It utilizes image features for soft prompt learning, and the prediction probability is computed as follows:
\begin{equation}
  p(c_k|x) = \frac {e^{cos(\mathcal{G}([cot(h^p_k), M(x), t_k]), \mathcal{H}(x)) / \tau}}{\sum_{i=1}^K e^{cos(\mathcal{G}([cot(h^p_i), M(x), t_i]), \mathcal{H}(x))/ \tau}},
\end{equation}
where $cot$ is a CoT module for prompt learning.

\subsection{Aggregation-Graph-of-Thought}
\label{sec:GoT}

The CoT achieves good results in various tasks due to its superior reasoning abilities. However, it cannot gather information from different instances in multiple views under a single reasoning step. This limitation restricts further assistance of reasoning in multi-modal understanding. 
To address this problem, we leverage the aggregation mechanism in the graph neural network and employ a chain of subgraphs $G =\{G_1, \dots, G_Z\}$ to construct the AGoT.
Each subgraph $G_i = \left(N_i, E_i\right)$ is a directed weighted graph built on a group of subnodes $n_i^s = \{n_{i1}^s, ..., n_{iR}^s\}$, which learns meta-prompts from $R$ different views and aggregates to the central node $N_i$ with the learned weights $E_i$ as edges, as shown in Figure~\ref{fig:framework}.
Further, the aggregated prompt representation is combined with the visual information extracted by the \textit{MetaNet} and then flows to the next subgraph with the flow controller.
After $Z$ reasoning steps, the final AGoT prompt representation $\mathcal{E}(G_Z)$ is generated and combined with the class label [CLASS] to build the final input prompt for the text encoder. Finally, a CLIP-like multi-modal contrastive learning network is adopted to make the classification or description of the image with the highest similarity.

\paragraph{Prompt Aggregation}
\label{subsec:prompt aggregation}
To better explore the prompt information from different perspectives, we propose a prompt aggregation approach for each reasoning step of the AGoT to generate high-quality prompt representation. Detailly, in each subgraph $G_i$, $R$ Gaussian-initialized meta-prompt learners act as subnodes $n_i^s$, and \textit{WeightNet} learns the aggregation edge weights $E_i$ from the image representation $\mathcal{H}(x)$. The central node prompt representation $\mathcal{E}(N_i)$ can be expressed as:
\begin{equation}
    \label{eq:prompt_agg}
    \begin{aligned}
        \mathcal{E}(N_i) &= \sum_{r=1}^R \mathcal{E}(n_{ir}^s) * E_{ir} \\ 
        where\ E_{ir} &= \textit{\textrm{WeightNet}}_{ir}(\mathcal{H}(x)) \\
        subject&~to~\sum_{r=1}^R E_{ir} = 1,
    \end{aligned}
\end{equation}
where $\textit{\textrm{WeightNet}}_{ir}(\cdot)$ is implemented as 3-layer MLP network with ReLU as activation function for subgraph $G_i$.%

As discussed in CoCoOp, it adds a \textit{MetaNet} network to encode the visual features generated by the image encoder and adds to the text prompt as a deviation term. 
We follow this design and introduce a \textit{MetaNet} network to aggregate the visual information into the central node representation. This aggregation process can be expressed as:
\begin{equation}
\label{eq:prompt_agg_meta}
    \mathcal{E}(G_i) = \mathcal{E}(N_i) + \textit{\textrm{MetaNet}}_i(\mathcal{H}(x)).
\end{equation}

\paragraph{Prompt Flow}

In AGoT, we decompose problem-solving into a step-by-step reasoning process, where each step relays the information from the previous steps to gradually deepen understanding.  
Intuitively, as the complexity of the task increases, a more intricate chain of reasoning graph becomes necessary.
To cope with this issue, we introduce a dynamic prompt information flow controller.
Specifically, given an input image, the encoded image features are fed to the flow controller, which is also a 3-layer MLP network with ReLU activation that generates an $\alpha$ to control the prompt fusion ratio between steps. The prompt flow control operation can be represented as:
\begin{equation}
\begin{aligned}
\label{eq:prompt_agg_meta_flow}
    \alpha_i & = \textit{\textrm{FlowControl}}_i(\mathcal{H}(x)), \\
    \mathcal{E}(G_i) &= (1 - \alpha_i) * \mathcal{E}(G_{i-1}) + \alpha_i * \mathcal{E}(G_i). \\
\end{aligned}
\end{equation}
 
After the $Z$ reasoning steps, the final AGoT prompt representation is defined as the last subgraph representation $\mathcal{E}(G_Z)$, which is combined with the image class label ``[CLASS]'' to build the input prompt for the text encoder. Finally, we conduct a standard CLIP-like multi-modal contrastive learning to train the AGoT network. To provide a more detailed description of the entire model's process, we use pseudocode to describe AGoT as shown in Algorithm~\ref{alg:AGoT}.
\begin{algorithm}
\renewcommand{\algorithmicrequire}{\textbf{Input:}}
\renewcommand{\algorithmicensure}{\textbf{Output:}}
\caption{Aggregation-Graph-of-Thought Algorithm}
\label{alg:AGoT}
\SetAlgoLined
\begin{algorithmic}[1]
    \REQUIRE $G =\{G_1, \dots, G_Z\}$: a chain of $Z$ subgraphs with initialized parameter $\Theta$ \\
    $n_i^s = \{n_{i1}^s, ..., n_{iR}^s\}$: a group of $R$ meta-prompt learners as subnodes for each $G_i$\\
    \textit{WeightNet}: the network that controls the aggregation weights of meta-prompts in $G_i$ \\
    \textit{FlowControl}: the network that controls the fusion ratio between steps in flow operation \\
    \textit{MetaNet}: the network that encodes the visual information as a deviation term \\
    $D$: the training dataset \\ 
    $T$: the training iterations \\
    \ENSURE Optimized $\Theta$ \\
    \FOR{$t=1,2,...,T$}
    \STATE Sample a batch of images $x$ from $D$
    \STATE Extract the image features $\mathcal{H}(x)$ with CLIP image encoder
    \STATE Compute the soft-prompt representation with AGoT
        \FOR{each subgraph $G_i$}
        \STATE Compute the central node representation $\mathcal{E}(N_i)$ with Eq.~\ref{eq:prompt_agg}
        \STATE Combine the visual information with Eq.~\ref{eq:prompt_agg_meta}
        \STATE Combine the prompts from the last step with Eq.~\ref{eq:prompt_agg_meta_flow}
        \ENDFOR
    \STATE Get the final AGoT representation $\mathcal{E}(G_Z)$
    \STATE Build the final input prompt to CLIP text encoder with $\mathcal{E}(G_Z)$ and [CLASS]
    \STATE Update $\Theta$ with CLIP-like multi-modal contrastive learning
    \ENDFOR
    \RETURN Optimized $\Theta$
\end{algorithmic}
\end{algorithm}

\section{Experiments}

We followed the exact experimental setup of previous works~\cite{zhou2022learning,zhou2022conditional} and validated the multi-modal reasoning performance of AGoT against recent state-of-the-art prompt learning models on widely-used benchmarks under various settings. Our approach was mainly evaluated in the following five primary tasks including text-image retrieval, VQA, cross label generalization, cross dataset generalization and domain generalization. Extensive experiments on a total of 18 datasets validate the effectiveness of our architecture. Meanwhile, we devised four ablation experiments and investigated the best practice of our method on various datasets.

\subsection{Setup}

\paragraph{Datasets}

For the text-image retrieval task, we use Flickr30k, which contains 29K/1K/1K for train/validation/test, and MSCOCO, which contains 113K/5K/5K for train/validation/test. For the VQA task, we use the VQAv2 dataset, which contains 1080K/210K/104K images paired with questions and answers for training/validation/test.

For the cross-label generalization ability of our proposed AGoT, we adopt 11 classification datasets, encompassing 2 object classification datasets (ImageNet~\cite{deng2009imagenet}, Caltech101~\cite{fei2004learning}), 5 fine-grained image recognition datasets (OxfordPets~\cite{parkhi2012cats}, StanfordCars~\cite{krause20133d}, Flowers102~\cite{nilsback2008automated}, Food101~\cite{bossard2014food}, FGVCAircraft~\cite{maji2013fine}), 1 scene recognition datasets (SUN397~\cite{xiao2010sun}), 1 texture classification datasets (DTD~\cite{cimpoi2014describing}), 1 satellite image classification datasets (EuroSAT~\cite{helber2019eurosat}), and an action classification (UCF101~\cite{soomro2012ucf101}).
For the cross-dataset generalization setting, all models are trained on ImageNet(Ima) in the source domain, and then the trained models are tested on Caltech101(Cal), OxfordPets(Oxf), StanfordCars(Sta), Flowers102(Flo), Food101(Foo), FGVCAircraft(FGV), SUN397(SUN), DTD, EuroSAT(Eur), and UCF101(UCF) in the target domain. 

For the domain generalization setting, we use ImageNet as the source domain dataset and evaluate performance on ImageNetV2~\cite{recht2019imagenet}, ImageNet-Sketch~\cite{wang2019learning}, ImageNet-A~\cite{hendrycks2021natural}, and ImageNet-R~\cite{hendrycks2021many}. Text and image classification were used for cross-dataset transfer evaluation and domain generalization evaluation. 
\paragraph{Evaluation Metrics}

For all tasks, we report the average result over three different random seeds. For the text-image retrieval and VQA tasks, we evaluate using recall at 1 (R@1) as a metric\footnote{Note that we follow \cite{ge2023chain} and consider VQA as a classification problem where the model selects the answer from a set of candidate answers based on the given question.}. In the base-to-new text-and-image classification setting, we report the harmonic mean $H=2\times(\text{base}\times\text{new})/(\text{base}+\text{new})$, which measures the generalization trade-off between the base and new sets. Finally, for the cross-dataset transfer and domain generalization setting, we use accuracy as the evaluation metric.

\paragraph{Training Details}

Consistent with CoCoOp, we chose CLIP with the vision backbone ViT-B/16 and loaded the pre-trained weights of CLIP, keeping them frozen during training. In the image-text retrieval and VQA tasks, for training the model, we employed image captions as class labels, set the learning rate to 0.02. The code can be obtained from \url{https://github.com/shishicode/AGoT}. For the three experiments on image classification, We utilized a chain length of 5 and initialized each of them with the phrase \textit{``a photo of a'' \{class\}} using a pre-trained CLIP text encoder. We used the class label as the class description and a learning rate of 0.002 to train the model and sample 16 shots from the base classes. 

\paragraph{Baselines}

We compare our proposed approach with the following state-of-the-art models: zero-shot CLIP with the fixed handcrafted prompt \textit{``a photo of a'' \{class\}}, CoOp, CoCoOp, ProGrad, KgCoOp, and CoT-PT.

\subsection{Results and Analysis}

\begin{table}[t]
\small
\centering
\footnotesize
\begin{tabular}{cccc} 
\toprule
Training data &Method &Flickr30k &MSCOCO\\
\midrule
0\% & CLIP & 83.00 & 53.30  \\ \midrule
\multirow{3}{*}{0.5\%} & CoCoOp & 82.80 & 53.50  \\
&CoT-PT  & 83.50 & 55.80  \\ 
&\textbf{AGoT} & \textbf{84.30} & \textbf{57.10}  \\ 
\midrule
\multirow{3}{*}{1.0\%} & CoCoOp & 84.50 & 56.40  \\
&CoT-PT & 85.10 & 56.70  \\ 
&\textbf{AGoT} & \textbf{87.60} & \textbf{57.60}  \\
\midrule
\multirow{3}{*}{2.0\%} & CoCoOp& 85.00 & 57.00  \\
&CoT-PT & 86.00 & 57.90  \\
&\textbf{AGoT} & \textbf{88.70} & \textbf{58.70}  \\
\bottomrule
\end{tabular}
\caption{Comparison between CLIP, CoCoOp, CoT-PT, and our AGoT methods on the Flickr30k and MSCOCO datasets.}
\label{lab:image-text-retrieval}
\end{table}

\begin{table}
\small
\centering
\footnotesize
\setlength\tabcolsep{16pt}
\begin{tabular}{ccc} 
\toprule
Training data &Method &VQAv2\\
\midrule
0\% & CLIP &  11.83  \\ \midrule
\multirow{3}{*}{0.25\%} & CoCoOp & 27.23\\
&CoT-PT & 29.13  \\
&\textbf{AGoT} & \textbf{29.64}  \\
\midrule
\multirow{3}{*}{0.5\%} & CoCoOp & 29.51\\
&CoT-PT & 30.72  \\ 
&\textbf{AGoT} & \textbf{31.31}  \\ 
\midrule
\multirow{3}{*}{0.75\%} & CoCoOp & 30.76\\
&CoT-PT & 30.86   \\ 
&\textbf{AGoT} & \textbf{31.74}   \\ 
\bottomrule
\end{tabular}
\caption{Comparison between CLIP, CoCoOp, CoT-PT, and our AGoT on the VQAv2 dataset.}\label{lab:visual-question-answer}
\vspace{-1.2em}
\end{table}

\begin{table*}[t]
\vspace{-0.8em}
\begin{minipage}{0.32\textwidth}
\centering
\footnotesize
 \subcaption{Average over 11 datasets}
\begin{tabular}{l|cc|c}

\toprule
	&Base & New & H\\
\midrule
CLIP 		& 69.34 & 74.22 & 71.70\\
CoOp		& \textbf{82.63} & 67.99 & 74.60\\
CoCoOp 	    & 80.47	& 71.69 & 75.83\\
ProGrad 	& 82.48& 70.75 & 76.16\\
KgCoOp		&80.73	& 73.60	& 77.00\\
CoT-PT		&80.23	& 74.20	& 77.10\\
\midrule
\textbf{AGoT}		&80.50	&\textbf{75.05} 	&\textbf{77.68} \\
\bottomrule
\end{tabular}
\end{minipage}
\hfill
\begin{minipage}{0.32\textwidth}
\centering
\footnotesize
 \subcaption{ImageNet}
\begin{tabular}{l|cc|c}
\toprule
	&Base & New & H\\
\midrule
CLIP 		&72.43 & 68.14 & 70.22\\
CoOp		&76.46 & 66.31 & 71.02\\
CoCoOp 	&75.98 & 70.43 & 73.10 \\
ProGrad 	&\textbf{77.02}	& 66.66 & 71.46\\
KgCoOp		&75.83 & 69.96	& 72.78\\
CoT-PT		&76.00	& 70.68	& 73.24\\
\midrule
\textbf{AGoT}		&76.02 & \textbf{70.80}	& \textbf{73.32}\\
\bottomrule
\end{tabular}
\end{minipage}
\hfill
\begin{minipage}{0.32\textwidth}
\centering
\footnotesize
\subcaption{Caltech101}
\begin{tabular}{l|cc|c}
\toprule
	&Base & New & H\\
\midrule
CLIP 		&96.84 &94.00 &95.40 \\
CoOp		&\textbf{98.11} &93.52 &95.76\\
CoCoOp 	&97.96 &93.81 &95.84 \\
ProGrad 	&98.02 & 93.89 & 95.91\\
KgCoOp		&97.72 & 94.39 & 96.03\\
CoT-PT		&97.91	& 94.03	& 95.93\\
\midrule
\textbf{AGoT}		&98.06 & \textbf{94.92} & \textbf{96.46}\\
\bottomrule
\end{tabular}
\end{minipage}
\hfill
\begin{minipage}{0.32\textwidth}
\centering
\footnotesize
 \subcaption{OxfordPets}
\begin{tabular}{l|cc|c}
\toprule
	&Base & New & H\\
\midrule
CLIP 		&91.17 &97.26 &94.12\\
CoOp		&94.24&96.66	&95.43\\
CoCoOp 	    &95.20 &97.69	&96.43 \\
ProGrad 	&95.07	&97.63  & 96.33 \\
KgCoOp		&94.65 &97.76   &96.18\\
CoT-PT		&\textbf{95.43}	& 97.78	& \textbf{96.59}\\
\midrule
\textbf{AGoT}		&94.94 &\textbf{98.19} &96.54\\
\bottomrule
\end{tabular}
\end{minipage}
\hfill
\begin{minipage}{0.32\textwidth}
\centering
\footnotesize
 \subcaption{StanfordCars}
\begin{tabular}{l|cc|c}
\toprule
	&Base & New & H\\
\midrule
CLIP 		&63.37 &74.89 &68.65\\
CoOp		&76.20 &69.14 &72.49\\
CoCoOp 	&70.49 &73.59 &72.01\\
ProGrad 	&\textbf{77.68} &68.63 &72.88\\
KgCoOp		&71.76 &\textbf{75.04} & 73.36\\
CoT-PT		&70.59	& 73.82	& 72.17\\
\midrule
\textbf{AGoT}		&72.23 &74.77& \textbf{73.47\textbf}\\
\bottomrule
\end{tabular}
\end{minipage}
\hfill
\begin{minipage}{0.32\textwidth}
\centering
\footnotesize
\subcaption{Flowers102}
\begin{tabular}{l|cc|c}
\toprule
	&Base & New & H\\
\midrule
CLIP 		&72.08 &\textbf{77.80} &74.83 \\
CoOp		&\textbf{97.63} &69.55 & 81.23\\
CoCoOp 	    &94.87&71.75&81.71\\
ProGrad 	&95.54&71.87&82.03\\
KgCoOp		&95.00&74.73&\textbf{83.65}\\
CoT-PT	    &94.46	& 72.46	& 82.01\\
\midrule
\textbf{AGoT}		&94.01 &74.18 &82.93\\
\bottomrule
\end{tabular}
\end{minipage}
\hfill
\begin{minipage}{0.32\textwidth}
\centering
\footnotesize
 \subcaption{Food101}
\begin{tabular}{l|cc|c}
\toprule
	&Base & New & H\\
\midrule
CLIP 		&90.10&91.22&90.66	\\
CoOp		&89.44&87.50&88.46	\\
CoCoOp 	&90.70&91.29&90.99\\
ProGrad 	&90.37&89.59& 89.98\\
KgCoOp		&90.50&91.70&91.09\\
CoT-PT		&90.74	& 91.77	& 91.25\\
\midrule
\textbf{AGoT}		&\textbf{90.88} &\textbf{92.81}&\textbf{91.83}\\
\bottomrule
\end{tabular}
\end{minipage}
\hfill
\begin{minipage}{0.32\textwidth}
\centering
\footnotesize
 \subcaption{FGVCAircraft}
\begin{tabular}{l|cc|c}
\toprule
	&Base & New & H\\
\midrule
CLIP 		&27.19&36.29&31.09	\\
CoOp		&39.24&30.49&34.30	\\
CoCoOp 	&33.41&23.71&	27.74\\
ProGrad 	&\textbf{40.54} &27.57& 32.82\\
KgCoOp		&36.21&33.55&\textbf{34.83}\\
CoT-PT		&35.13	& 32.21	&33.61\\
\midrule
\textbf{AGoT}		&35.23 &\textbf{34.14} &34.68\\
\bottomrule
\end{tabular}
\end{minipage}
\hfill
\begin{minipage}{0.32\textwidth}
\centering
\footnotesize
\subcaption{SUN397}
\begin{tabular}{l|cc|c}
\toprule
	&Base & New & H\\
\midrule
CLIP 		&69.36&75.35&	72.23\\
CoOp		&80.85&68.34&74.07\\
CoCoOp 	&79.74&76.86&78.27\\
ProGrad 	&\textbf{81.26}&74.17& 77.55\\
KgCoOp		&80.29& 76.53& 78.36\\
CoT-PT		&79.44	& 77.20	& 78.30\\
\midrule
\textbf{AGoT}		&79.59& \textbf{77.60}& \textbf{78.58}\\
\bottomrule
\end{tabular}
\end{minipage}
\hfill
\begin{minipage}{0.32\textwidth}
\centering
\footnotesize
 \subcaption{DTD}
\begin{tabular}{l|cc|c}
\toprule
	&Base & New & H\\
\midrule
CLIP 		&53.24&\textbf{59.90}&56.37\\
CoOp		&\textbf{80.17}&47.54&59.68\\
CoCoOp 	    &77.01&56.00 & 64.85\\
ProGrad 	&77.35&52.35&62.45\\
KgCoOp		&77.55& 54.99& 64.35\\
CoT-PT		&76.27	& 58.34	& \textbf{66.11}\\
\midrule
\textbf{AGoT}		&76.73& 57.88& 65.99\\
\bottomrule
\end{tabular}
\end{minipage}
\hfill
\begin{minipage}{0.32\textwidth}
\centering
\footnotesize
 \subcaption{EuroSAT}
\begin{tabular}{l|cc|c}
\toprule
	&Base & New & H\\
\midrule
CLIP 		&56.48&64.05&60.03\\
CoOp		&91.54&54.44&68.27\\
CoCoOp 	    &87.49&60.04&71.21	\\
ProGrad 	&\textbf{90.11}&60.89&72.67\\
KgCoOp		&85.64& 64.34& 73.48\\
CoT-PT		&84.11	& \textbf{72.81}	& 78.06\\
\midrule
\textbf{AGoT}		&86.11& 71.67& \textbf{78.23}\\
\bottomrule
\end{tabular}
\end{minipage}
\hfill
\begin{minipage}{0.32\textwidth}
\centering
\footnotesize
\subcaption{UCF101}
\begin{tabular}{l|cc|c}
\toprule
	&Base & New & H\\
\midrule
CLIP 		&70.53&77.50&73.85\\
CoOp		&\textbf{85.14}&64.47&	73.37\\
CoCoOp 	&82.33&73.45&77.64	\\
ProGrad 	&84.33&74.94&79.35\\
KgCoOp		&82.89& 76.67& 79.65\\
CoT-PT		&82.47	& 75.09	& 78.61\\
\midrule
\textbf{AGoT}		&81.90 &\textbf{78.68}&\textbf{80.25} \\
\bottomrule
\end{tabular}
\end{minipage}
\caption{Comparison with existing methods (CLIP, CoOp, CoCoOp, ProGrad, KgCoOp, CoT-PT) in the cross-label generalization setting with ViT-B/16 as the backbone. 
}
\label{tab:base2new}
\vspace{-1.0em}
\end{table*}

\paragraph{Text-Image Retrieval}

\begin{table*}[t]
    \tabstyle{5pt}
    \small
    \vspace{-0.8em}
    \begin{tabular}{l c ccccccccccc}
    \toprule
    & Source & \multicolumn{11}{c}{Target} \\ \cmidrule(lr){2-2} \cmidrule(lr){3-13}
    & Ima & Cal & Oxf & Sta & Flo & Foo & FGV & SUN & DTD & Eur & UCF  &Avg \\
     \midrule
    CoOp & 71.51 & 93.70 & 89.14 & 64.51 & 68.71 & 85.30 & 18.47 & 64.15 & 41.92 & 46.39 & 66.55 & 63.88 \\
    CoCoOp & 71.02 & 94.43 & 90.14 & 65.32 & 71.88 & 86.06 & 22.94 & 67.36 & 45.73 & 45.37 & 68.21 & 65.74 \\
    ProGrad & 70.21 & 94.43 & 93.21 & 71.75 & 89.98 & 85.77 & \textbf{32.93} & 71.17 & 57.72 & {70.84} & 77.82 & 74.21 \\
    KgCoOp & 70.19 & 94.65 & 93.20 & 71.98 & \textbf{90.69} & 86.59 & 32.47 & 71.79 & 58.31 & {71.06} & 78.40 & 74.48 \\
    
    \midrule
    \textbf{AGoT} & \textbf{72.12} & \textbf{95.92} & \textbf{98.25} & \textbf{75.11} & 75.73 & \textbf{91.20} & 32.01 & \textbf{76.53} & \textbf{58.75} & \textbf{71.46} & \textbf{79.48} & \textbf{75.44} \\
    \bottomrule
    \end{tabular}
    \caption{Comparison of CoOp, CoCoOp, ProGrad, KgCoOp and our AGoT methods in the cross dataset generalization setting.}
    \label{tab:cross dataset}
\end{table*}

\begin{table*}[]
\centering
\small
\vspace{-0.8em}
\begin{tabular}{l|c|c|cccc|c}
\toprule
    &\multirow{2}{*}{Prompts}& Source   & \multicolumn{5}{c}{Target}                             \\\cline{3-8} 
    & & ImageNet & ImageNetV2 & ImageNet-Sketch & ImageNet-A & ImageNet-R & Avg. \\
\midrule
CLIP &  hand-crafted	& 66.73    	& 60.83    	& 46.15          	& 47.77      	& 73.96      & 57.17\\ 
UPT & vp+tp & \textbf{72.63} & 64.35 & 48.66 & 50.66 & 76.24 & 59.98 \\
CoCoOp & vp+tp  	& 71.02    	& 64.07      	& 48.75           	& 50.63      	& 76.18      & 59.90 \\
CoOp &  tp  	& 71.51    	& 64.20       	& 47.99           	& 49.71      	& 75.21    & 59.28   \\
ProGrad & tp 	& 72.24    	& \textbf{64.73}      	& 47.61           	& 49.39      	& 74.58   & 59.07   \\
KgCoOp   &  tp & 71.20    	& 64.10       	& 48.97           	& \textbf{50.69}     	& 76.70  & \textbf{60.11}  \\
\midrule
\textbf{AGoT}   &  vp+tp & 72.12    	& 64.54       	&\textbf{49.76}            	& 50.26     	& \textbf{76.92}  & \textbf{60.37}  \\
\bottomrule
\end{tabular}
\caption{Comparison of CLIP, UPT, CoCoOp, CoOp, ProGrad, KgCoOp and AGoT in the domain generalization setting. Where ``vp'' and ``tp'' denote the visual prompting and textual prompting, respectively.}
\label{tab:domain generalization}
\vspace{-0.8em}
\end{table*}

\begin{table}
\scalebox{0.92}{
\footnotesize
\setlength\tabcolsep{3pt}
\centering
    \begin{tabular}{ccccccc}
    \toprule
    \small
    Sample ratio      &  0.2\%  &  0.5\% &1.0\% &2.0\% &5.0\%  &8.0\%  \\ 
        \midrule
    steps=3        & 83.30 & 84.30 &87.60 &88.70 &89.50 &\textbf{90.10}  \\ 
        \midrule
    steps=5        & 83.20 & 84.50 &87.10 &88.40 &89.60 &\textbf{90.30}  \\
        \bottomrule   
    \end{tabular}
    }
    \caption{Relationship between sample ratio and reasoning steps on the Flickr30k dataset.}
 \label{lib:step-radio}
\end{table}

\begin{table}
\footnotesize
\setlength\tabcolsep{6pt}
\centering
    \begin{tabular}{cccccc}
    \toprule
    \small
        Number & 2 & 3 & 4 & 5 & 6 \\ \midrule
        Recall   & 86.90  & 88.20 & 87.90 & \textbf{88.70} & 88.20 \\ \bottomrule
    \end{tabular}
   \caption{Comparison the effect of the number of aggregation subnodes on the Flickr30k dataset.}
   \label{lib:aggregation-number}
   \vspace{-0.8em}
\end{table}

\begin{table}
\footnotesize
\setlength\tabcolsep{6pt}
\centering
    \begin{tabular}{cccccc}
    \toprule
    \small
        $\alpha$   & 0.1    & 0.3   & 0.5   & 0.7   & Dynamic \\ 
        \midrule
        Recall   & 57.80  & 58.00 & 58.10 & 57.90 & \textbf{58.70} \\ \bottomrule
    \end{tabular}
    \caption{Comparison of the effect of flow controller with the fixed and dynamic prompt fusion ratios on the MSCOCO dataset.}
 \label{lib:flow-controller}
\end{table}

In this experiment, we evaluated our model on two image captioning datasets: Flickr30k and MSCOCO. We trained and tested our model using settings with 0.5\%, 1.0\%, and 2.0\% of the training data and evaluated it on the test set. As can be observed from Table~\ref{lab:image-text-retrieval}, our method outperformed the CLIP, CoCoOp, and CoT-PT in all tests. On both the Flickr30k and MSCOCO datasets, when the training data consisted of only 0.5\% of the total data, AGoT performed 0.80\% and 1.30\% better than the current SOTA method CoT-PT; when the training data was 1.0\% of the total data, AGoT performed 2.5\% and 0.9\% better than CoT-PT; and when the training data was 2.0\% of the total data, AGoT performed 1.7\% and 0.8\% better than CoT-PT, respectively. These improvements demonstrate that our proposed AGoT is more effective for multi-modal representation learning than CoT, which was the previous state-of-the-art. Additionally, our proposed novel soft-prompting is helpful for cross-modal representation learning without tuning the pre-training parameters of CLIP.

\begin{table*}
\scalebox{0.94}{
\footnotesize
\setlength\tabcolsep{4pt}
\centering
    \begin{tabular}{cccccccccccc}
    \toprule
    \small
        Subnodes & 1 & 2 & $\Delta$ & 3 & $\Delta$ & 4 & $\Delta$ & 5 & $\Delta$ & 6 & $\Delta$ \\ \midrule
        Recall   & 86.00  & 86.90 & +0.90 & 88.20 & +2.20 &87.90 &+1.90 &\textbf{88.70} &+2.70 &88.20 &+2.20 \\ \midrule
        Parameters(M)   & 149.74  & 149.76 & +0.02 & 149.78 & +0.04 &149.81 &+0.07 &149.83 &+0.09 &149.85 &+0.11 \\ \midrule
        Reasoning Time(s)   & 180  & 182 & +2 & 187 & +7 &190 & +10 &191 & +11 &193 &+13 \\ \midrule
        Training Time(s) &208.11 &216.42 &+8.31 &222.57 & +14.46 & 227.34 &+19.23 &231.24 & +23.13 & 235.63 &+27.52\\
        \bottomrule
    \end{tabular}
    }
   \caption{Evaluation of the impact on Recall, Parameters, Reasoning time, and Training time with increasing aggregation subnodes on the Flickr30k dataset.}
   \label{lib:multiple-view}
   \vspace{-0.8em}
\end{table*}

\paragraph{VQA}

In VQA, we also sample 0.25\%, 0.5\%, and 0.75\% from the training set following~\cite{ge2023chain}. The results are shown in Table~\ref{lab:visual-question-answer}. Similar to image-to-text retrieval, our method outperforms CLIP, CoCoOp, and CoT-PT in all settings. AGoT obtains 0.51\%, 0.59\%, and 0.88\% gains over CoT-PT, respectively. Experiments show that our AGoT approach can also benefit relatively more complex VQA tasks due to better reasoning and multi-modal representation learning. 

\paragraph{Cross Label Generalization}

In addition to being evaluated on multi-modal understanding tasks, based on the conclusion of previous work~\cite{zhou2022conditional}, better prompting methods can lead to better domain generalization abilities. Therefore, we first conducted experiments on the cross-label generalization setting.
Specifically, we conducted experiments using AGoT on 11 datasets, where each dataset was divided into base and new classes to represent the source and target domains. To train the model, we considered 16 samples per class and set the reasoning step $Z = 5$. The results, summarized in Table~\ref{tab:base2new}, revealed that AGoT outperformed other models such as CLIP, CoOp, CoCoOp, ProGrad, KgCoOp, and CoT-PT. Specifically, AGoT achieved accuracy improvements of 5.98\%, 3.08\%, 1.85\%, 1.52\%, 0.68\%, and 0.58\% over these strong baselines, respectively.
These experiments demonstrate that AGoT exhibits strong generalization performance when encountering new data, which results from multiple aspects of thought reasoning.

\paragraph{Cross Dataset Generalization}

In addition to cross-label generalization, we also explore the cross-dataset generalization setting. As shown in Table~\ref{tab:cross dataset}, AGoT outperforms the CoOp by 0.61\% on ImageNet (source). Furthermore, when applying the model to the target domain of 10 datasets, our results demonstrate a 0.96\% improvement over previous state-of-the-art KgCoOp. These experimental findings also provide solid evidence of AGoT's robust transferability compared to ordinary CoT. 

\paragraph{Domain Generalization}

Table~\ref{tab:domain generalization} shows the results of domain generalization. Our model trained on ImageNet was tested on four other ImageNet datasets of different types, and our method outperforms CLIP, UPT, CoCoOp, CoOp, ProGrad, KgCoOp. This confirms that CoT does have a good effect on domain generalization, and AGoT is an effective improvement of CoT, bringing stronger reasoning generalization ability.

\subsection{Ablation Study}

\paragraph{Reasoning Steps}

\begin{figure}[]
\center{\includegraphics[width=7.5cm]{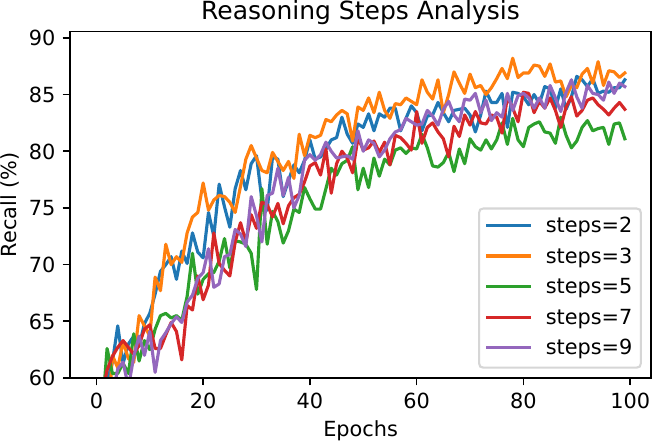}} 
\vspace{-0.8em}
   \caption{Comparison of different reasoning steps on the Flickr30k dataset.}
   \label{fig:length-analysis}
   \vspace{-1.2em}
\end{figure}

To demonstrate the impact of reasoning steps in AGoT, we conducted experiments on the image-text retrieval task using the Flickr30K dataset with a training ratio of 2.0\%. Specifically, we varied the reasoning steps from 2 to 9. The results, as shown in Figure ~\ref{fig:length-analysis}, indicate that AGoT achieves its best performance when the epoch is set to 80 and the reasoning steps are set to 3. This can be attributed to the fact that when the data size is small, longer reasoning steps in AGoT may be unstable and susceptible to overfitting. Thus, increasing the reasoning steps does not always lead to improved performance.
 
As shown in Table~\ref{lib:step-radio}, continuing to increase the proportion of training samples, when the training sample proportion reaches 5.0\%, the performance of inference step 5 surpasses that of inference step 3, achieving optimal performance. Similarly, the same conclusion is drawn when the sample proportion increases to 8.0\%. This indicates that as the dataset grows relatively larger and tasks become more complex, AGoT with longer inference steps can explore and embed more valuable features.

\paragraph{Number of Aggregation Subnodes}

To demonstrate the impact of the number of subnodes (i.e., prompts) aggregated by AGoT on reasoning performance, we conducted experiments on the Flickr30K dataset using a train data ratio of 2.0\%. Specifically, we varied the number of subnodes from 2 to 6. As shown in Table~\ref{lib:aggregation-number}, we found that the best result was achieved when the number of aggregations was set to 5. This suggests that having more aspect prompts tends to improve performance, but optimal performance depends on task complexity.

\paragraph{Role of Flow Controller}

In the AGoT model, we use a flow controller to dynamically regulate the degree of message passing between central nodes. To evaluate the effectiveness of this flow controller, we contrasted it with the fixed prompt fusion ratios $\alpha$ of 0.1, 0.3, 0.5, and 0.7 on the MSCOCO dataset using 2.0\% sampled data. And the reasoing steps and subnodes are fixed to 3 and 4, respectively. The results, presented in Table~\ref{lib:flow-controller}, demonstrate that the dynamic flow controller outperforms the fixed ratios. This improvement can be attributed to the varying features of different images and their requirement for distinct reasoning processes. By employing input-based dynamic control, the model achieves better reasoning and enhances its overall performance.

\paragraph{Role of Multi-View Thinking}
To provide a more comprehensive analysis of the impact of the number of aggregation subnodes on the computational and model complexity of AGoT, we conduct experiments on the Flickr30k dataset with a sample proportion of 2.0\%.
In these experiments, the number of subnodes is systematically varied from 1 to 6, we find that the best Recall is achieved when the number of aggregations is set to 5. Notably, our method degenerates to COT-PT when the number of subnodes is 1. 
The outcomes of these experiments are detailed in Table~\ref{lib:multiple-view}. 

When increasing aggregation subnodes from 1 to 2, Recall improved by +0.90\%, parameters increased by +0.02M, reasoning time increased by +2s, and training time increased by +8.31s. Similarly, when increasing aggregation subnodes from 1 to 5, Recall improved by +2.70\%, parameters increased by +0.09M, reasoning time increased by +11s, and training time increased by +23.13s. Hence, it is evident that with the increasing number of aggregation subnodes, the model's parameters, reasoning time and training time exhibit minimal changes to achieve a favorable trade-off. Please refer to Supplementary Material Section~\ref{app:other_abla} for the impact of different key values, different backbone architectures, and flow controllers on the model across other datasets.

\section{Conclusion}
In this paper, we introduce a novel AGoT mechanism for soft-prompt tuning in multi-modal representation learning. The mechanism consists of a novel prompt aggregation and prompt flow process designed to improve multi-modal reasoning. We demonstrate the effectiveness of this method through experiments conducted on 2 text-image retrieval tasks, 1 VQA task, and 15 image and text classification tasks. In future work, we plan to introduce a reasoning early exiting mechanism to address the varying complexity of inputs, which may require different reasoning strategies.

\section*{Acknowledgements}
This research was supported by the National Natural Science Foundation of China (No. 62306216, 62376200), the Natural Science Foundation of Hubei Province
of China (No. 2023AFB816), the Fundamental Research Funds for
the Central Universities (No. 2042023kf0133).

\section*{Limitations}
Applying AGoT in small models under fine-tuning scenarios. Typically, limited by capacity, a small model requires fine-tuning to acquire thinking abilities. A joint training schema may intensify the problem's complexity. The stability of CoT-based prompt technology currently lacks consistency and is sensitive to hyper-parameters. Designing the prompt structure, determining its steps, and establishing proper initialization methods all require extensive experimentation and exploration. Furthermore, as the reasoning steps increase, the overall speed of the system decreases significantly, and there is a higher consumption of GPU memory.

\section*{Ethics Statement}
The mechanism presented in this paper is a robust tool with the potential to enhance multi-modal representation learning and improve system generalization. However, society needs to maintain a steadfast focus on data privacy and remain cautious about the potential for technology to be maliciously misused. 


\section*{References}


\bibliographystyle{lrec-coling2024-natbib}
\bibliography{lrec-coling2024-example}

\begin{thebibliography}{49}
\expandafter\ifx\csname natexlab\endcsname\relax\def\natexlab#1{#1}\fi

\bibitem[{Afouras et~al.(2018)Afouras, Chung, Senior, Vinyals, and Zisserman}]{afouras2018deep}
Triantafyllos Afouras, Joon~Son Chung, Andrew Senior, Oriol Vinyals, and Andrew Zisserman. 2018.
\newblock Deep audio-visual speech recognition.
\newblock \emph{IEEE transactions on pattern analysis and machine intelligence}, 44(12):8717--8727.

\bibitem[{Bossard et~al.(2014)Bossard, Guillaumin, and Van~Gool}]{bossard2014food}
Lukas Bossard, Matthieu Guillaumin, and Luc Van~Gool. 2014.
\newblock Food-101--mining discriminative components with random forests.
\newblock In \emph{Computer Vision--ECCV 2014: 13th European Conference, Zurich, Switzerland, September 6-12, 2014, Proceedings, Part VI 13}, pages 446--461. Springer.

\bibitem[{Chen et~al.(2015)Chen, Fang, Lin, Vedantam, Gupta, Doll{\'a}r, and Zitnick}]{chen2015microsoft}
Xinlei Chen, Hao Fang, Tsung-Yi Lin, Ramakrishna Vedantam, Saurabh Gupta, Piotr Doll{\'a}r, and C~Lawrence Zitnick. 2015.
\newblock Microsoft coco captions: Data collection and evaluation server.
\newblock \emph{arXiv preprint arXiv:1504.00325}.

\bibitem[{Cimpoi et~al.(2014)Cimpoi, Maji, Kokkinos, Mohamed, and Vedaldi}]{cimpoi2014describing}
Mircea Cimpoi, Subhransu Maji, Iasonas Kokkinos, Sammy Mohamed, and Andrea Vedaldi. 2014.
\newblock Describing textures in the wild.
\newblock In \emph{Proceedings of the IEEE conference on computer vision and pattern recognition}, pages 3606--3613.

\bibitem[{Deng et~al.(2009)Deng, Dong, Socher, Li, Li, and Fei-Fei}]{deng2009imagenet}
Jia Deng, Wei Dong, Richard Socher, Li-Jia Li, Kai Li, and Li~Fei-Fei. 2009.
\newblock Imagenet: A large-scale hierarchical image database.
\newblock In \emph{2009 IEEE conference on computer vision and pattern recognition}, pages 248--255. Ieee.

\bibitem[{Dosovitskiy et~al.(2021)Dosovitskiy, Beyer, Kolesnikov, Weissenborn, Zhai, Unterthiner, Dehghani, Minderer, Heigold, Gelly, Uszkoreit, and Houlsby}]{DBLP:conf/iclr/DosovitskiyB0WZ21}
Alexey Dosovitskiy, Lucas Beyer, Alexander Kolesnikov, Dirk Weissenborn, Xiaohua Zhai, Thomas Unterthiner, Mostafa Dehghani, Matthias Minderer, Georg Heigold, Sylvain Gelly, Jakob Uszkoreit, and Neil Houlsby. 2021.
\newblock \href {https://openreview.net/forum?id=YicbFdNTTy} {An image is worth 16x16 words: Transformers for image recognition at scale}.
\newblock In \emph{9th International Conference on Learning Representations, {ICLR} 2021, Virtual Event, Austria, May 3-7, 2021}. OpenReview.net.

\bibitem[{Fei-Fei et~al.(2004)Fei-Fei, Fergus, and Perona}]{fei2004learning}
Li~Fei-Fei, Rob Fergus, and Pietro Perona. 2004.
\newblock Learning generative visual models from few training examples: An incremental bayesian approach tested on 101 object categories.
\newblock In \emph{2004 conference on computer vision and pattern recognition workshop}, pages 178--178. IEEE.

\bibitem[{Gabeur et~al.(2022)Gabeur, Nagrani, Sun, Alahari, and Schmid}]{gabeur2022masking}
Valentin Gabeur, Arsha Nagrani, Chen Sun, Karteek Alahari, and Cordelia Schmid. 2022.
\newblock Masking modalities for cross-modal video retrieval.
\newblock In \emph{Proceedings of the IEEE/CVF Winter Conference on Applications of Computer Vision}, pages 1766--1775.

\bibitem[{Ge et~al.(2023)Ge, Luo, Qian, Gan, Fu, and Zhan}]{ge2023chain}
Jiaxin Ge, Hongyin Luo, Siyuan Qian, Yulu Gan, Jie Fu, and Shanghang Zhan. 2023.
\newblock Chain of thought prompt tuning in vision language models.
\newblock \emph{arXiv preprint arXiv:2304.07919}.

\bibitem[{Gu et~al.(2022)Gu, Han, Liu, and Huang}]{DBLP:conf/acl/GuHLH22}
Yuxian Gu, Xu~Han, Zhiyuan Liu, and Minlie Huang. 2022.
\newblock \href {https://doi.org/10.18653/V1/2022.ACL-LONG.576} {{PPT:} pre-trained prompt tuning for few-shot learning}.
\newblock In \emph{Proceedings of the 60th Annual Meeting of the Association for Computational Linguistics (Volume 1: Long Papers), {ACL} 2022, Dublin, Ireland, May 22-27, 2022}, pages 8410--8423. Association for Computational Linguistics.

\bibitem[{Han et~al.(2022)Han, Zhao, Ding, Liu, and Sun}]{DBLP:journals/aiopen/HanZDLS22}
Xu~Han, Weilin Zhao, Ning Ding, Zhiyuan Liu, and Maosong Sun. 2022.
\newblock \href {https://doi.org/10.1016/j.aiopen.2022.11.003} {{PTR:} prompt tuning with rules for text classification}.
\newblock \emph{{AI} Open}, 3:182--192.

\bibitem[{He et~al.(2016)He, Zhang, Ren, and Sun}]{he2016deep}
Kaiming He, Xiangyu Zhang, Shaoqing Ren, and Jian Sun. 2016.
\newblock Deep residual learning for image recognition.
\newblock In \emph{Proceedings of the IEEE conference on computer vision and pattern recognition}, pages 770--778.

\bibitem[{He et~al.(2023)He, Li, Cai, and Wang}]{he2023multi}
Liqi He, Zuchao Li, Xiantao Cai, and Ping Wang. 2023.
\newblock Multi-modal latent space learning for chain-of-thought reasoning in language models.
\newblock \emph{arXiv preprint arXiv:2312.08762}.

\bibitem[{Helber et~al.(2019)Helber, Bischke, Dengel, and Borth}]{helber2019eurosat}
Patrick Helber, Benjamin Bischke, Andreas Dengel, and Damian Borth. 2019.
\newblock Eurosat: A novel dataset and deep learning benchmark for land use and land cover classification.
\newblock \emph{IEEE Journal of Selected Topics in Applied Earth Observations and Remote Sensing}, 12(7):2217--2226.

\bibitem[{Hendrycks et~al.(2021{\natexlab{a}})Hendrycks, Basart, Mu, Kadavath, Wang, Dorundo, Desai, Zhu, Parajuli, Guo et~al.}]{hendrycks2021many}
Dan Hendrycks, Steven Basart, Norman Mu, Saurav Kadavath, Frank Wang, Evan Dorundo, Rahul Desai, Tyler Zhu, Samyak Parajuli, Mike Guo, et~al. 2021{\natexlab{a}}.
\newblock The many faces of robustness: A critical analysis of out-of-distribution generalization.
\newblock In \emph{Proceedings of the IEEE/CVF International Conference on Computer Vision}, pages 8340--8349.

\bibitem[{Hendrycks et~al.(2021{\natexlab{b}})Hendrycks, Zhao, Basart, Steinhardt, and Song}]{hendrycks2021natural}
Dan Hendrycks, Kevin Zhao, Steven Basart, Jacob Steinhardt, and Dawn Song. 2021{\natexlab{b}}.
\newblock Natural adversarial examples.
\newblock In \emph{Proceedings of the IEEE/CVF Conference on Computer Vision and Pattern Recognition}, pages 15262--15271.

\bibitem[{Jia et~al.(2021)Jia, Yang, Xia, Chen, Parekh, Pham, Le, Sung, Li, and Duerig}]{jia2021scaling}
Chao Jia, Yinfei Yang, Ye~Xia, Yi-Ting Chen, Zarana Parekh, Hieu Pham, Quoc Le, Yun-Hsuan Sung, Zhen Li, and Tom Duerig. 2021.
\newblock Scaling up visual and vision-language representation learning with noisy text supervision.
\newblock In \emph{International Conference on Machine Learning}, pages 4904--4916. PMLR.

\bibitem[{Krause et~al.(2013)Krause, Stark, Deng, and Fei-Fei}]{krause20133d}
Jonathan Krause, Michael Stark, Jia Deng, and Li~Fei-Fei. 2013.
\newblock 3d object representations for fine-grained categorization.
\newblock In \emph{Proceedings of the IEEE international conference on computer vision workshops}, pages 554--561.

\bibitem[{Li et~al.(2022)Li, Li, Li, Niebles, and Hoi}]{li2022align}
Dongxu Li, Junnan Li, Hongdong Li, Juan~Carlos Niebles, and Steven~CH Hoi. 2022.
\newblock Align and prompt: Video-and-language pre-training with entity prompts.
\newblock In \emph{Proceedings of the IEEE/CVF Conference on Computer Vision and Pattern Recognition}, pages 4953--4963.

\bibitem[{Liu et~al.(2022)Liu, Ji, Fu, Tam, Du, Yang, and Tang}]{liu2022p}
Xiao Liu, Kaixuan Ji, Yicheng Fu, Weng Tam, Zhengxiao Du, Zhilin Yang, and Jie Tang. 2022.
\newblock P-tuning: Prompt tuning can be comparable to fine-tuning across scales and tasks.
\newblock In \emph{Proceedings of the 60th Annual Meeting of the Association for Computational Linguistics (Volume 2: Short Papers)}, pages 61--68.

\bibitem[{Lu et~al.(2022)Lu, Liu, Zhang, Liu, and Tian}]{lu2022prompt}
Yuning Lu, Jianzhuang Liu, Yonggang Zhang, Yajing Liu, and Xinmei Tian. 2022.
\newblock Prompt distribution learning.
\newblock In \emph{Proceedings of the IEEE/CVF Conference on Computer Vision and Pattern Recognition}, pages 5206--5215.

\bibitem[{Maji et~al.(2013)Maji, Rahtu, Kannala, Blaschko, and Vedaldi}]{maji2013fine}
Subhransu Maji, Esa Rahtu, Juho Kannala, Matthew Blaschko, and Andrea Vedaldi. 2013.
\newblock Fine-grained visual classification of aircraft.
\newblock \emph{arXiv preprint arXiv:1306.5151}.

\bibitem[{Nilsback and Zisserman(2008)}]{nilsback2008automated}
Maria-Elena Nilsback and Andrew Zisserman. 2008.
\newblock Automated flower classification over a large number of classes.
\newblock In \emph{2008 Sixth Indian Conference on Computer Vision, Graphics \& Image Processing}, pages 722--729. IEEE.

\bibitem[{Parkhi et~al.(2012)Parkhi, Vedaldi, Zisserman, and Jawahar}]{parkhi2012cats}
Omkar~M Parkhi, Andrea Vedaldi, Andrew Zisserman, and CV~Jawahar. 2012.
\newblock Cats and dogs.
\newblock In \emph{2012 IEEE conference on computer vision and pattern recognition}, pages 3498--3505. IEEE.

\bibitem[{Plummer et~al.(2015)Plummer, Wang, Cervantes, Caicedo, Hockenmaier, and Lazebnik}]{plummer2015flickr30k}
Bryan~A Plummer, Liwei Wang, Chris~M Cervantes, Juan~C Caicedo, Julia Hockenmaier, and Svetlana Lazebnik. 2015.
\newblock Flickr30k entities: Collecting region-to-phrase correspondences for richer image-to-sentence models.
\newblock In \emph{Proceedings of the IEEE international conference on computer vision}, pages 2641--2649.

\bibitem[{Radford et~al.(2021)Radford, Kim, Hallacy, Ramesh, Goh, Agarwal, Sastry, Askell, Mishkin, Clark et~al.}]{radford2021learning}
Alec Radford, Jong~Wook Kim, Chris Hallacy, Aditya Ramesh, Gabriel Goh, Sandhini Agarwal, Girish Sastry, Amanda Askell, Pamela Mishkin, Jack Clark, et~al. 2021.
\newblock Learning transferable visual models from natural language supervision.
\newblock In \emph{ICML}.

\bibitem[{Recht et~al.(2019)Recht, Roelofs, Schmidt, and Shankar}]{recht2019imagenet}
Benjamin Recht, Rebecca Roelofs, Ludwig Schmidt, and Vaishaal Shankar. 2019.
\newblock Do imagenet classifiers generalize to imagenet?
\newblock In \emph{International Conference on Machine Learning}, pages 5389--5400. PMLR.

\bibitem[{Ren et~al.(2019)Ren, Ruan, Tan, Qin, Zhao, Zhao, and Liu}]{ren2019fastspeech}
Yi~Ren, Yangjun Ruan, Xu~Tan, Tao Qin, Sheng Zhao, Zhou Zhao, and Tie-Yan Liu. 2019.
\newblock Fastspeech: Fast, robust and controllable text to speech.
\newblock \emph{Advances in neural information processing systems}, 32.

\bibitem[{Rose et~al.(2023)Rose, Himakunthala, Ouyang, He, Mei, Lu, Saxon, Sonar, Mirza, and Wang}]{rose2023visual}
Daniel Rose, Vaishnavi Himakunthala, Andy Ouyang, Ryan He, Alex Mei, Yujie Lu, Michael Saxon, Chinmay Sonar, Diba Mirza, and William~Yang Wang. 2023.
\newblock Visual chain of thought: Bridging logical gaps with multimodal infillings.
\newblock \emph{arXiv preprint arXiv:2305.02317}.

\bibitem[{Schick and Sch{\"{u}}tze(2021)}]{DBLP:conf/eacl/SchickS21}
Timo Schick and Hinrich Sch{\"{u}}tze. 2021.
\newblock \href {https://doi.org/10.18653/V1/2021.EACL-MAIN.20} {Exploiting cloze-questions for few-shot text classification and natural language inference}.
\newblock In \emph{Proceedings of the 16th Conference of the European Chapter of the Association for Computational Linguistics: Main Volume, {EACL} 2021, Online, April 19 - 23, 2021}, pages 255--269. Association for Computational Linguistics.

\bibitem[{Shen et~al.(2022)Shen, Li, Tan, Bansal, Rohrbach, Chang, Yao, and Keutzer}]{DBLP:conf/iclr/ShenLTBRCYK22}
Sheng Shen, Liunian~Harold Li, Hao Tan, Mohit Bansal, Anna Rohrbach, Kai{-}Wei Chang, Zhewei Yao, and Kurt Keutzer. 2022.
\newblock \href {https://openreview.net/forum?id=zf\_Ll3HZWgy} {How much can {CLIP} benefit vision-and-language tasks?}
\newblock In \emph{The Tenth International Conference on Learning Representations, {ICLR} 2022, Virtual Event, April 25-29, 2022}. OpenReview.net.

\bibitem[{Shi et~al.(2023)Shi, Suzgun, Freitag, Wang, Srivats, Vosoughi, Chung, Tay, Ruder, Zhou, Das, and Wei}]{DBLP:conf/iclr/ShiSF0SVCTRZ0W23}
Freda Shi, Mirac Suzgun, Markus Freitag, Xuezhi Wang, Suraj Srivats, Soroush Vosoughi, Hyung~Won Chung, Yi~Tay, Sebastian Ruder, Denny Zhou, Dipanjan Das, and Jason Wei. 2023.
\newblock \href {https://openreview.net/pdf?id=fR3wGCk-IXp} {Language models are multilingual chain-of-thought reasoners}.
\newblock In \emph{The Eleventh International Conference on Learning Representations, {ICLR} 2023, Kigali, Rwanda, May 1-5, 2023}. OpenReview.net.

\bibitem[{Soomro et~al.(2012)Soomro, Zamir, and Shah}]{soomro2012ucf101}
Khurram Soomro, Amir~Roshan Zamir, and Mubarak Shah. 2012.
\newblock Ucf101: A dataset of 101 human actions classes from videos in the wild.
\newblock \emph{arXiv preprint arXiv:1212.0402}.

\bibitem[{Sun et~al.(2022)Sun, Hu, and Saenko}]{sun2022dualcoop}
Ximeng Sun, Ping Hu, and Kate Saenko. 2022.
\newblock Dualcoop: Fast adaptation to multi-label recognition with limited annotations.
\newblock \emph{Advances in Neural Information Processing Systems}, 35:30569--30582.

\bibitem[{Vaswani et~al.(2017)Vaswani, Shazeer, Parmar, Uszkoreit, Jones, Gomez, Kaiser, and Polosukhin}]{vaswani2017attention}
Ashish Vaswani, Noam Shazeer, Niki Parmar, Jakob Uszkoreit, Llion Jones, Aidan~N Gomez, {\L}ukasz Kaiser, and Illia Polosukhin. 2017.
\newblock Attention is all you need.
\newblock \emph{Advances in neural information processing systems}, 30.

\bibitem[{Wang et~al.(2019)Wang, Ge, Lipton, and Xing}]{wang2019learning}
Haohan Wang, Songwei Ge, Zachary Lipton, and Eric~P Xing. 2019.
\newblock Learning robust global representations by penalizing local predictive power.
\newblock \emph{Advances in Neural Information Processing Systems}, 32.

\bibitem[{Wang et~al.(2023)Wang, Wei, Schuurmans, Le, Chi, Narang, Chowdhery, and Zhou}]{DBLP:conf/iclr/0002WSLCNCZ23}
Xuezhi Wang, Jason Wei, Dale Schuurmans, Quoc~V. Le, Ed~H. Chi, Sharan Narang, Aakanksha Chowdhery, and Denny Zhou. 2023.
\newblock \href {https://openreview.net/pdf?id=1PL1NIMMrw} {Self-consistency improves chain of thought reasoning in language models}.
\newblock In \emph{The Eleventh International Conference on Learning Representations, {ICLR} 2023, Kigali, Rwanda, May 1-5, 2023}. OpenReview.net.

\bibitem[{Wei et~al.(2022)Wei, Wang, Schuurmans, Bosma, Xia, Chi, Le, Zhou et~al.}]{wei2022chain}
Jason Wei, Xuezhi Wang, Dale Schuurmans, Maarten Bosma, Fei Xia, Ed~Chi, Quoc~V Le, Denny Zhou, et~al. 2022.
\newblock Chain-of-thought prompting elicits reasoning in large language models.
\newblock \emph{Advances in neural information processing systems}, 35:24824--24837.

\bibitem[{Xiao et~al.(2010)Xiao, Hays, Ehinger, Oliva, and Torralba}]{xiao2010sun}
Jianxiong Xiao, James Hays, Krista~A Ehinger, Aude Oliva, and Antonio Torralba. 2010.
\newblock Sun database: Large-scale scene recognition from abbey to zoo.
\newblock In \emph{2010 IEEE computer society conference on computer vision and pattern recognition}, pages 3485--3492. IEEE.

\bibitem[{Yao et~al.(2023)Yao, Zhang, and Xu}]{yao2023visual}
Hantao Yao, Rui Zhang, and Changsheng Xu. 2023.
\newblock Visual-language prompt tuning with knowledge-guided context optimization.
\newblock In \emph{Proceedings of the IEEE/CVF Conference on Computer Vision and Pattern Recognition}, pages 6757--6767.

\bibitem[{Yao et~al.(2024)Yao, Yu, Zhao, Shafran, Griffiths, Cao, and Narasimhan}]{yao2024tree}
Shunyu Yao, Dian Yu, Jeffrey Zhao, Izhak Shafran, Tom Griffiths, Yuan Cao, and Karthik Narasimhan. 2024.
\newblock Tree of thoughts: Deliberate problem solving with large language models.
\newblock \emph{Advances in Neural Information Processing Systems}, 36.

\bibitem[{Yu et~al.(2022)Yu, Fei, and Li}]{yu2022u}
Tan Yu, Hongliang Fei, and Ping Li. 2022.
\newblock U-bert for fast and scalable text-image retrieval.
\newblock In \emph{Proceedings of the 2022 ACM SIGIR International Conference on Theory of Information Retrieval}, pages 193--203.

\bibitem[{Zelikman et~al.(2022)Zelikman, Mu, Goodman, and Wu}]{zelikman2022star}
Eric Zelikman, Jesse Mu, Noah~D Goodman, and Yuhuai~Tony Wu. 2022.
\newblock Star: Self-taught reasoner bootstrapping reasoning with reasoning.

\bibitem[{Zhang et~al.(2023{\natexlab{a}})Zhang, Zhang, Li, and Smola}]{DBLP:conf/iclr/0001Z0S23}
Zhuosheng Zhang, Aston Zhang, Mu~Li, and Alex Smola. 2023{\natexlab{a}}.
\newblock \href {https://openreview.net/pdf?id=5NTt8GFjUHkr} {Automatic chain of thought prompting in large language models}.
\newblock In \emph{The Eleventh International Conference on Learning Representations, {ICLR} 2023, Kigali, Rwanda, May 1-5, 2023}. OpenReview.net.

\bibitem[{Zhang et~al.(2023{\natexlab{b}})Zhang, Zhang, Li, Zhao, Karypis, and Smola}]{zhang2023multimodal}
Zhuosheng Zhang, Aston Zhang, Mu~Li, Hai Zhao, George Karypis, and Alex Smola. 2023{\natexlab{b}}.
\newblock Multimodal chain-of-thought reasoning in language models.
\newblock \emph{arXiv preprint arXiv:2302.00923}.

\bibitem[{Zheng et~al.(2021)Zheng, Yin, Chen, Ma, Liu, and Yang}]{zheng2021knowledge}
Wenfeng Zheng, Lirong Yin, Xiaobing Chen, Zhiyang Ma, Shan Liu, and Bo~Yang. 2021.
\newblock Knowledge base graph embedding module design for visual question answering model.
\newblock \emph{Pattern recognition}, 120:108153.

\bibitem[{Zhou et~al.(2022{\natexlab{a}})Zhou, Yang, Loy, and Liu}]{zhou2022conditional}
Kaiyang Zhou, Jingkang Yang, Chen~Change Loy, and Ziwei Liu. 2022{\natexlab{a}}.
\newblock Conditional prompt learning for vision-language models.
\newblock In \emph{Proceedings of the IEEE/CVF Conference on Computer Vision and Pattern Recognition}, pages 16816--16825.

\bibitem[{Zhou et~al.(2022{\natexlab{b}})Zhou, Yang, Loy, and Liu}]{zhou2022learning}
Kaiyang Zhou, Jingkang Yang, Chen~Change Loy, and Ziwei Liu. 2022{\natexlab{b}}.
\newblock Learning to prompt for vision-language models.
\newblock \emph{International Journal of Computer Vision}, 130(9):2337--2348.

\bibitem[{Zhu et~al.(2023)Zhu, Niu, Han, Wu, and Zhang}]{zhu2023prompt}
Beier Zhu, Yulei Niu, Yucheng Han, Yue Wu, and Hanwang Zhang. 2023.
\newblock Prompt-aligned gradient for prompt tuning.
\newblock In \emph{Proceedings of the IEEE/CVF International Conference on Computer Vision}, pages 15659--15669.

\end{thebibliography}

\bibliographystylelanguageresource{lrec-coling2024-natbib}
\bibliographylanguageresource{languageresource}

\appendix
\section*{Appendix}

\section{Ablation Study}
\label{app:other_abla}
\paragraph{Different Key Value}
To investigate the influence of different sample quantities on model performance, we selected three values for single category quantity: 4, 8, and 16, across 11 classification task datasets. We compared the model's training on the Base domain and testing on the New domain. The results were contrasted with those of AGoT model against CoOp, CoCoOp, ProGrad, and KgCoOp, as detailed in Table~\ref{lib:differ_key_value}. When K is 4, AGoT achieved a value of 76.87\%, surpassing CoOp by +4.43\% and KgCoOp by +0.97\%. When K is 8, AGoT reached a value of 76.74\%, outperforming CoOp by +3.24\% and KgCoOp by +0.68\%. When K is 16, AGoT achieved a value of 77.68\%, surpassing CoOp by +3.08\% and KgCoOp by +0.68\%.
\begin{table*}[]

    \centering
    \vspace{-0.8em}
    \begin{tabular}{ccccccccccc}
    \toprule
        \multirow{2}{*}{Backbone} & \multirow{2}{*}{Methods}   & \multicolumn{3}{c}{K=4}   & \multicolumn{3}{c}{K=8} & \multicolumn{3}{c}{K=16}  \\ \cline{3-11}
        &  & {Base} & {New} & {H} &  {Base} & {New} & {H} &  {Base} & {New} & {H} \\
    \hline

    \multirow{5}{*}{ViT-B/16} & CoOp   	& 78.43    	       	& 68.03           	& 72.44 & \textbf{80.73}    	       	& 68.39           	& 73.50   & \textbf{82.63}    	       	& 67.99           	& 74.60      \\
    & CoCoOp &76.72 &73.34 &74.85 &78.56 &72.00 &74.90 &80.47 &71.69 &75.83   \\
    & ProGrad  	 &79.18 &71.14 &74.62 &80.62 &71.02 &75.20 &82.48 &70.75 &76.16 \\
    & KgCoOp       &\textbf{79.92} &73.11 &75.90 &78.36 &73.89 &76.06 &80.73 &73.60 &77.00   \\
    &\textbf{AGoT}    &79.34 &\textbf{74.56} &\textbf{76.87} &78.19 &\textbf{75.34} &\textbf{76.74} &80.50 &\textbf{75.05} &\textbf{77.68} \\
    \bottomrule
    \hline
    \end{tabular}

\caption{Comparing the effects of different K values on model performance}
\label{lib:differ_key_value}
\vspace{-0.8em}
\end{table*}

\paragraph{Different Backbone}
To assess the impact of different backbone networks on model performance, we conducted tests on four networks: ViT-B/16, ViT-B/32, ResNet-50, and ResNet-101. The experimental results are shown in Table~\ref{lib:diff_backbone}. From Table~\ref{lib:diff_backbone}, it can be observed that the model performs best when ViT-B/16 is used as the backbone. Therefore, the choice of different backbones has a significant impact on the model's performance.

\begin{table*}
\footnotesize
\centering
    \begin{tabular}{ccccccccccc}
    \toprule
    \small
        Backbones & Cal & Oxf & Stan & Flo & Foo & FGV & DTD & Eur & UCF & Ave \\ \midrule
        ViT-B/16   & \textbf{96.46}  & \textbf{96.54} & \textbf{73.47} & \textbf{82.93} & \textbf{91.83} & \textbf{34.68} & \textbf{65.99} &\textbf{78.23} &\textbf{80.25} &\textbf{77.82} \\ \midrule
        ViT-B/32   & 95.32  & 94.93 & 67.42 & 74.37 & 87.46 & 26.93 & 61.43 &71.46 &76.58 &72.88 \\ \midrule
        ResNet-50  & 93.61  & 94.37 & 64.39 & 76.18 & 85.19 & 24.64 & 58.36 &49.28 &72.39 &68.71 \\ \midrule
        ResNet-101  & 94.57  & 94.16 & 68.54 & 76.34 & 87.42 & 8.67 & 61.87 &54.54 &76.86 &69.22 \\ 
        \bottomrule
    \end{tabular}
   \caption{Comparative analysis of the impact of different backbones on model performance.}
   \label{lib:diff_backbone}
   \vspace{-0.8em}
\end{table*}

\paragraph{Flow controller on other datasets}

To further validate the impact of dynamic and fixed values of the flow controller on the nine datasets, As shown in Table~\ref{lib:diff_flow_contro}, it is evident that dynamic values are beneficial for improving the model's performance.
\begin{table*}
\footnotesize
\centering
    \begin{tabular}{ccccccccccc}
    \toprule
    \small
        Backbones & Cal & Oxf & Stan & Flo & Foo & FGV & DTD & Eur & UCF & Ave \\ \midrule
        ViT-B/16(dynamic)   & \textbf{96.46}  & \textbf{96.54} & \textbf{73.47} & \textbf{82.93} & \textbf{91.83} & \textbf{34.68} & \textbf{65.99} &\textbf{78.23} &\textbf{80.25} &\textbf{77.82} \\ \midrule
        ViT-B/16(fix=0.5)   & 95.72  & 95.81 & 72.36 & 81.37 & 90.64 & 33.93 & 64.61 &77.45 &79.48 &76.79 \\ 
        \bottomrule
    \end{tabular}
   \caption{Impact of fixed and dynamic flow controller on model performance}
   \label{lib:diff_flow_contro}
   \vspace{-0.8em}
\end{table*}

\end{document}